\pdfoutput=1

\documentclass[11pt]{article}

\usepackage[preprint]{acl}

\usepackage{times}
\usepackage{latexsym}

\usepackage[T1]{fontenc}

\usepackage[utf8]{inputenc}

\usepackage{microtype}
\usepackage{amsmath}
\usepackage{amsfonts}

\usepackage{inconsolata}

\usepackage{graphicx}

\usepackage{booktabs}

\definecolor{light_blue}{HTML}{DCE6F1}
\usepackage[most]{tcolorbox}
\tcbset{on line, 
        boxsep=1pt, left=0pt,right=0pt,top=0pt,bottom=0pt,
        colframe=white,colback=light_blue,  
        highlight math style={enhanced}
        }

\title{Language Models Encode Numbers Using Digit Representations in Base 10}

\author{Amit Arnold Levy\thanks{\hspace{3px}Work done at Tel Aviv University.} \\
  University of Oxford \\
  \texttt{amit.levy@keble.ox.ac.uk} \\\And
  Mor Geva \\
  Tel Aviv University \\
  \texttt{morgeva@tauex.tau.ac.il} \\}

\begin{document}
\maketitle

\begin{abstract}
Large language models (LLMs) frequently make errors when handling even simple numerical problems, such as comparing two small numbers. 
A natural hypothesis is that these errors stem from how LLMs represent numbers, and specifically, whether their representations of numbers capture their numeric values.
We tackle this question from the observation that LLM errors on numerical tasks are often distributed across \textit{the digits} of the answer rather than normally around \textit{its numeric value}. Through a series of probing experiments and causal interventions, we show that LLMs internally represent numbers with individual circular representations per-digit in base 10.
This digit-wise representation, as opposed to a value representation, sheds light on the error patterns of models on tasks involving numerical reasoning and could serve as a basis for future studies on analyzing numerical mechanisms in LLMs.
\end{abstract}

\section{Introduction}
Despite their high performance on various challenging tasks \citep{bubeck2023sparksartificialgeneralintelligence, bommasani2021opportunities, trinh2024solving}, large language models (LLMs) often struggle with simple numerical problems, such as adding or comparing the magnitude of two small numbers. 
While previous works commonly attribute such failures to different limitations in the representations of LLMs \citep[e.g.,][]{mcleish2024transformers, nogueira2021investigatinglimitationstransformerssimple}, \textit{how} LLMs represent numbers is still an outstanding question.

Recently, \citet{zhu2024languagemodelsknowvalue} used linear probes to predict the number encoded in a hidden representation, showing high correlation with the expected value. However, the probes exhibited low accuracy, suggesting that a linear representation alone is not sufficient to explain how LLMs can often perform exact numerical operations, such as addition and multiplication. \citet{Maltoni_2024} have suggested that LLMs may do arithmetic in ``value space'', but then we would expect to see a normally-distributed error pattern, which we will see is not the case in widely-used models.

\begin{figure}[t]
\centering
\includegraphics[scale=0.46]{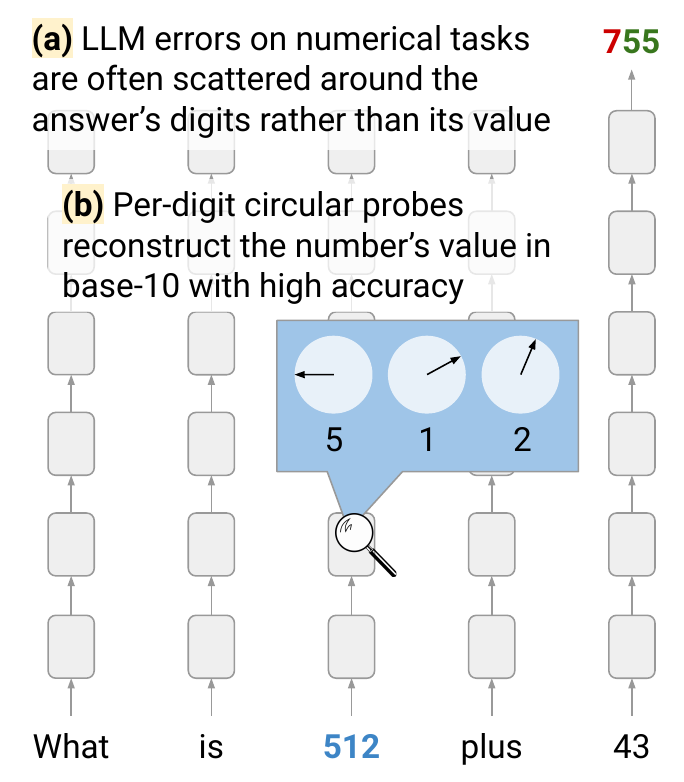}
\caption{An illustration of our key findings, suggesting that LLMs represent numbers on a per-digit base-10 basis: (a) on simple numerical tasks, LLMs often make errors that are close to the answer in `digit space' rather than in value space, (b) though probing the exact number is hard, digit values can be decoded accurately.}
\label{fig:intro}
\end{figure}

We approach the above question by observing that when models make numerical errors, the errors are often distant from the correct answer in value space but close in `digit space'.
For example, consider the simple addition problem  ${\text{``}132+238+324+139=\text{''}}$ where the correct answer is 833. LLMs are more likely to generate errors with high string-similarity to the correct answer, such as ``633'' or ``823'', than natural errors like ``831'' or ``834'', which are close in value,
as if the model's internal algorithm misreads one of the digits in the input. We show this rigorously in \S\ref{sec:error_distribution}.

We argue that such scattered error distributions are unlikely to occur in models that directly manipulate numbers in a value space. For example, in multi-operand addition, if the model represents each number in a value space and then translates the result back to tokens after addition, we would expect a normal error distribution around the correct answer. This distribution would arise from noise in the addition operation and the representations themselves. However, the observed scattered errors (in \S\ref{sec:error_distribution}) suggest that the model may represent numbers in a fragmented manner, for example based on their individual digits.

To test this hypothesis, we first train probes to recover the number value and digit values from hidden representations of numbers. 
Our experiments with Llama 3 8B \citep{dubey2024llama3herdmodels} and Mistral 7B \citep{jiang2023mistral7b} show that, while probes fail to recover the exact number value directly \citep[which agrees with][]{zhu2024languagemodelsknowvalue},
the hidden representations of a number contain an orthogonal circular representation for each digit in base 10 (as illustrated in Figure~\ref{fig:intro}).
This observation holds across both models, which use different tokenization schemes for numbers.
Moreover, causally intervening on these circular digit representations (i.e., performing $+5$ mod10) often modifies the value of the number accurately.

\begin{figure}[t]
\setlength\belowcaptionskip{-10pt}
\centering
   \includegraphics[scale=0.33]{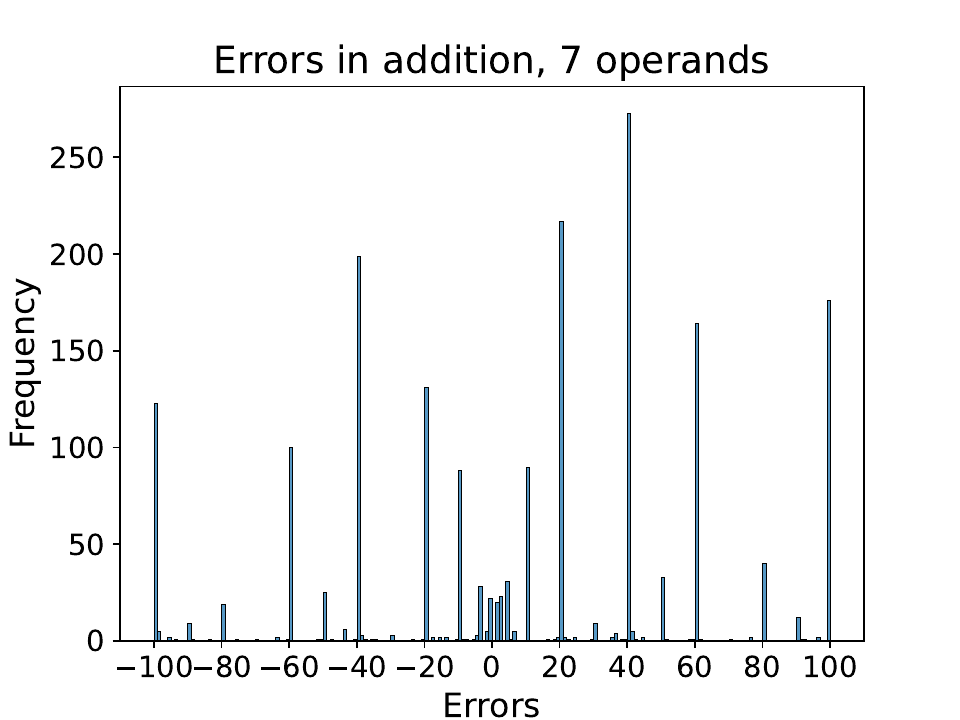}
  \caption{Error distribution in 7 operand addition.}
    \label{fig:7operand}
\end{figure}

To conclude, our work proposes that the scattered errors LLMs demonstrate on arithmetic problems stem from a fragmented digit-wise representation of numbers. We show that this hypothesis holds in practice; it is possible to accurately recover and modify the digit values from number representations in base 10, but not the number values. Our findings provide a basis for understanding mathematical operations in LLMs and mitigating numerical errors. We release our code at \url{https://github.com/amitlevy/base10}.

\section{Model Errors on Numerical Tasks are Scattered Across Digits}
\label{sec:error_distribution}

We analyze the distribution of errors by Llama 3 8B on two simple numerical tasks with numbers within the range $0$ to $999$, which the model represents as individual tokens. We find that errors are distributed in a digit-wise manner, where an incorrect prediction is close to the correct answer in string edit distance but not in value space. 

\paragraph{Task 1: Multi-operand addition} 
We generated 5,000 queries of addition of $N=7$ operands, which sum into a number between 0 and 1000, and calculated the errors of the model on these queries.
Figure~\ref{fig:7operand} displays the error distribution, showing that most errors are exact multiples of 10 and 100. 
Further, when considering the error distributions for any number of operands between 4 and 8, we observe that about 80\% of the errors are in a single output digit, which is often not the units digit.
A similar error analysis of GPT-4o \citep{openai2024gpt4ocard} on 20-operand addition tasks showed similar trends (\S\ref{apx:gpt4o}).

\paragraph{Task 2: Comparison of two numbers} 
We consider all pairs of numbers between 0 and 999, which differ from each other in only a single digit—the units, tens, or hundreds place. Given a pair of numbers, the model needs to indicate which number is larger, e.g. \textit{``between 121 or 171, the larger number is:''}.
Table~\ref{table:singledigitcomp} shows that errors are distributed approximately equally between the digits. This indicates that the model's likelihood of making a mistake is not significantly affected by the numerical closeness of the numbers, as would be expected if numbers were represented in value space.

\begin{table}[t]
\centering
\footnotesize
\begin{tabular}{l*{2}{c}}
\toprule
Digit & Correct & Incorrect \\
\midrule
Units          & 4,232 \tcbox{\footnotesize{94\%}}   & 259 \tcbox{\footnotesize{6\%}} \\
Tens           & 4,351 \tcbox{\footnotesize{97\%}}   & 140 \tcbox{\footnotesize{3\%}} \\
Hundreds       & 4,054 \tcbox{\footnotesize{92\%}}   & 356 \tcbox{\footnotesize{8\%}} \\
\bottomrule
\end{tabular}
\caption{Accuracy of Llama 3 8B in comparing the magnitude of two numbers differing by one digit.}
\label{table:singledigitcomp}
\end{table}

The evident base-10 digit-related error trends in both tasks lead to the hypothesis that LLMs may represent numbers in base 10 as opposed to in a linear value space, which we test in the next section.\looseness-1

\section{LLMs Represent Numbers Digit-Wise in Base 10}
\label{sec:main_results}

We test our hypothesis and show that LLMs represent numbers on a per-digit base 10 basis.

\begin{table*}[t]
\setlength\tabcolsep{4pt}
\footnotesize
\centering
\begin{tabular}{l*{15}{l}}
\toprule
Basis &  2    &  3    &  4    &  5    &  6    &  7    &  8    &  9    &  10   &  11   &  12   &  13   &  14   & 1000 & 2000 \\
\midrule
Llama 3 8B   &  0.16 &  0.06 &  0.16 &  \underline{0.67} &  0.05 &  0.08 &  0.06 &  0.07 &  \textbf{0.91} &  0.08 &  0.06 &  0.06 &  0.06 & 0.00 & 0.00 \\
Mistral 7B &  0.13 &  0.02 &  0.13 &  \underline{0.72} &  0.02 &  0.05 &  0.05 &  0.08 &  \textbf{0.92} &  0.12 &  0.04 &  0.06 &  0.05 & 0.01 & 0.00 \\
\bottomrule
\end{tabular}
\caption{Accuracy in predicting all digits of digit-wise circular probes in various bases, averaged over layers $\geq$ 3.}
\label{table:accuracy}
\end{table*}

\subsection{Experimental setting}

\paragraph{Probing} 
We train digit-wise probes that estimate the value of a number from its hidden representation by predicting the numeric values of its digits.
Let $M$ be a pre-trained transformer-based language model \citep{vaswani2023attentionneed} with $L$ layers and a hidden dimension $d$, and denote by $\mathbf{h}^\ell_{j}$ the hidden representation of the $j$-th input token at layer $\ell$. In the following, we omit the position index and use $\mathbf{h}^\ell$, as in our experiments we always consider the last position of the input (i.e., the last numeric token).
For a digit $i$, a base $b$, and a layer $\ell \in [L]$, we train a circular probe \citep{engels2024languagemodelfeatureslinear} that given the hidden representation $\mathbf{h}^\ell$ of a number $x$, predicts the numeric value of its $i$-th digit in base $b$:\looseness-1
{\begin{equation}
\label{eq:circ_probe}
\mathbf{P}_{i,b}^\ell = \underset{{\mathbf{P}'\in\mathbb{R}^{2\times d}}}{\arg\min} \sum_{\langle \mathbf{h}^\ell, x_i \rangle \in \mathcal{D}^\ell} \left\|\mathbf{P}' \mathbf{h^\ell} - \texttt{circle}_b(x_i) \right\|_2^2 
\end{equation}}
$\mathcal{D}^\ell$ is a training set consisting of pairs $\langle \mathbf{h}^\ell, x_i \rangle$ of the $\ell$-th layer hidden representation and the $i$-th digit of a number $x$, and
\begin{equation}
\label{eq:circle}
    \texttt{circle}_b(t) = [\cos(2\pi t/b),\sin(2\pi t/b)]
\end{equation} 
maps a digit in base $b$ to a point on the unit circle.

Using the set of probes for some layer $\ell$, we define a function that reconstructs the value of a number $x$ from its representation $\mathbf{h}^\ell$.
For every digit $i$, define a function ${\texttt{digit}_{i,b}^\ell: \mathbb{R}^d \to [b]}$
that predicts the value of that digit by applying $\texttt{digit}_{i,b}^\ell := \frac{b}{2\pi} \cdot \text{atan2}(\mathbf{P}_{i,b}^\ell \mathbf{h}^\ell)$.\footnote{atan2 computes the two argument arctangent, which we convert from a signed to an unsigned angle between 0 and $2\pi$.} Concatenating the outputs of the functions for all the digits of $x$ provides an estimation of its value in base $b$. For example, the value of a 3-digit number would be reconstructed in base $b$ from its $\ell$-layer representation by concatenating  $[ \texttt{digit}_{3,b}^\ell, \texttt{digit}_{2,b}^\ell, \texttt{digit}_{1,b}^\ell ]$.

In addition to the circular probes, we trained linear probes, which have been used recently to extract various features from LLM representations \citep{belinkov-2022-probing, park2023the, gurnee2024language}. While the linear probes showed similar trends to the circular probes, we observed they are less effective in predicting numerical values from LLM representations. This observation agrees with recent findings that some features in LLMs have non-linear representations \citep{engels2024languagemodelfeatureslinear} as well as with the circular patterns observed in PCA plots (see \S\ref{apx:pca}). Therefore, in our experiments we focus on circular probes.

\paragraph{Data} 
For each positive number $x\in[2000]$ we feed $\text{``}\langle x\rangle\text{''}$ (the value of $x$ as a string) as input to the model and extracted the hidden representations from every layer $\ell\in [L]$. In cases when $x$ is tokenized into multiple tokens, we take the representation at the last position (we assume that $M$ is an auto-regressive model).
For each basis $b$, we randomly split the numbers into train and validation sets with 1800 and 200 numbers, respectively. 

\paragraph{Models}
We analyze two popular auto-regressive decoder-only LLMs: Llama 3 8B \citep{dubey2024llama3herdmodels} and Mistral 7B \citep{jiang2023mistral7b}. Llama's tokenizer contains individual tokens for all numbers between 0 and 999 inclusive, which is the common choice for modern LLMs (e.g., GPT-4 \citet{singh2024tokenizationcountsimpacttokenization} and Claude Sonnet 3.5). Mistral 7B was picked for having a different tokenization from Llama, specifically a single token per digit, which can be expected to impose a stronger bias towards digit-wise representations of numbers.

\subsection{Probing recovers digit values in base 10 but not the whole number value}

Table~\ref{table:accuracy} shows the average probe accuracy over layers $\geq 3$ in predicting all the digits of the number correctly (maximum accuracy results show similar trends; see \S\ref{apx:probes}). We do not consider the early layers as multi-token numbers require multiple layers to contextualize (see Figure~\ref{fig:accuracy_per_layer} in \S\ref{apx:probes}).

The highest accuracy of 0.91 for Llama 3 8B and 0.92 for Mistral 7B is achieved when reconstructing the numbers in base 10. Moreover, for all other bases, accuracy is substantially lower, typically not exceeding 0.2, serving as a natural baseline for the base 10 results.
Specifically, classifying the number directly (base 2000) succeeds in only $<1\%$ of the cases, which further shows that the direct circular representation of the value in the hidden space is not accurate enough for arithmetic, similarly to the linear representation mentioned earlier. Interestingly, base 5 also has relatively high accuracy, though significantly below base 10.

Overall, these results show that while reconstructing the number value directly generally fails, reconstructing digit-by-digit in base 10 succeeds with high accuracy. 
Importantly, while such a representation has advantages (see discussion in \S\ref{sec:discussion}), it is surprising considering that LLMs typically have individual tokens for multi-digit numbers, which is not naturally base 10. We show evidence that the probes extend to representations of word form numbers, without being trained on them, in \S\ref{apx:wordform}.

\subsection{Modifying a digit representation modifies the whole number value accordingly}
\label{sec:intervention}

Our experiments suggest that models may represent numbers in a per-digit base 10 basis rather than store the number value directly.
Here we conduct a causal experiment to test if this digit-wise representation is used by Llama 3 8B during inference.

\begin{figure}[t]
\centering
\includegraphics[scale=0.42]{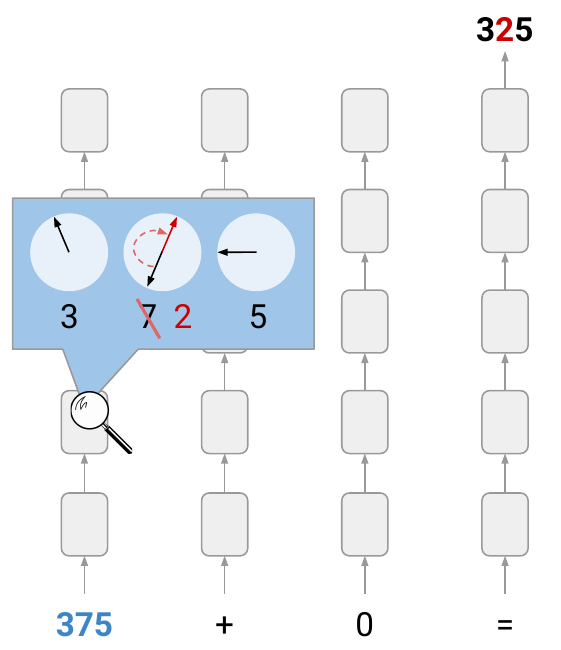}
\caption{An illustration of our intervention on number representations via circular per-digit probes in base 10.}
\label{fig:intervention}
\end{figure}

\paragraph{Experiment}
Since the digit representations are circular in base 10, if we flip a number's hidden representation along the two directions of the probe (Eq.~\ref{eq:circle}), we would expect the modified representation to encode the same number but with one digit flipped, i.e. the digit corresponding to the probe will now take a value of $v+5\ (\text{mod } 10)$ where $v$ was the original digit value before the intervention.
For example (Figure~\ref{fig:intervention}), flipping the tens digit in the representation of $375$ is expected to produce a representation of $325$.
For more details see \S\ref{apx:causaldetails}.

To test this intervention, we consider the model's inference pass on a query $\text{``}\langle x\rangle + 0=\text{''}$ with some number $x$, for which the model initially generates $x$ as the output. Then, we intervene on the representation of $x$ at layer $\ell$, apply the procedure described above to change one of $x$'s digits, and continue the model's run to obtain a new output $x'$.
Let $x_i$ and $x'_i$ be the $i$-th digits of $x$ and $x'$, we then check whether $x'_i = x_i + 5\ (\text{mod } 10)$ and for all $j\neq i$ that $x'_j=x_j$. 
We further define the prediction to be ``close'' to the intended result if it is closer to the intended result than an off by 1 error in the intervention digit. 
We conduct this experiment using all natural numbers 0 through 999. For each number, we perform the intervention once for every digit at layer 3, where the probes extract the number with high accuracy and before the information would propagate to the last position from which the prediction is obtained.

\paragraph{Results}
For the hundreds digit, the exact intended result was achieved 15\% of the time, while 47\% of the results were 'close' to the intended number, e.g., the digit was changed but with an error of 1 from the intended outcome. These numbers were respectively 10\% and 50\% for the tens digit, and 15\% and 50\% for the units digit.
As a baseline, using a linear intervention following \citet{zhu2024languagemodelsknowvalue}, but with the appropriate change to the normalization such that a specific number is targeted instead of a general direction, the exact result is achieved in less than 1\% of the cases. A random baseline would be replacing the numeric token with another random numeric token in the range of the intervention, leading to a random baseline accuracy of 0.1\%.

We conclude that there is a causal significance to the digit-wise circular representation, but there might be secondary representations or that some information might transfer before layer $3$.

\section{Related Work}
\label{sec:related_work}

\paragraph{Representation of numbers in LLMs} There has been some investigation into how LLMs may represent numeric magnitude, involving linear probes of hidden representations \citep{zhu2024languagemodelsknowvalue,heinzerling-inui-2024-monotonic} and embeddings \citep{wallace2019nlpmodelsknownumbers}. In \citet{gould2023successorheadsrecurringinterpretable} the authors looked into modular features of the first layer's hidden representations, and observed that modulus 10 seems of particular importance, but did not look beyond the units digit. To the best of our knowledge, no prior work has succeeded in training probes that extract the value of a held-out number from an LLM representation with the precision necessary to explain LLMs' successes on arithmetic.

\paragraph{Mechanistic interpretability of arithmetic tasks} There has been much interest in looking into how LLMs may perform arithmetic tasks. Recent work has largely focused on either performing in depth analysis of the algorithms learned by toy models \citep{Maltoni_2024, nanda2023progressmeasuresgrokkingmechanistic, quirke2024understandingadditiontransformers, yehudai2024can} or analyzing information flow in trained open source LLMs \citep{stolfo2023mechanisticinterpretationarithmeticreasoning, chen2024stateshiddenhiddenstates}. 
Most recently, \citet{zhou2024pretrained} demonstrated that LLMs utilize Fourier features for arithmetic operations, with distinct roles for low- and high-frequency components. Our work complements these efforts and provides a basis for future work in this avenue, by analyzing the representations of numbers in modern LLMs.

\paragraph{Failures of LLMs on arithmetic tasks}
\citet{razeghi-etal-2022-impact} have looked into the performance of GPT-J 6B on arithmetic tasks, showing it is correlated with the frequency of the terms in the training dataset, which potentially suggests that LLMs may not be reasoning at all.
While explaining LLM errors with number frequencies is valuable and may be more plausible in terms of the performance seen in older models, Llama 3 8B can perform 7 operand, 2-digit addition ($10^{14}$ possible problems) with about $50\%$ accuracy, which is far beyond the number of problems that could possibly be in the training data.

\section{Conclusion and Discussion}
\label{sec:discussion}

While previous research has demonstrated that linear probes struggle to accurately extract numerical values from hidden representations --- which are necessary for performing exact arithmetic operations like addition and multiplication --- our findings indicate that circular digit-wise probes can effectively achieve this in two models with different tokenization.
We have further demonstrated that editing these representations can alter the encoded number and consequently the model generation. 
These nonlinear representations align with \citet{engels2024languagemodelfeatureslinear}, who showed circular representations for the days of the week and months of the year.

\paragraph{Why would models construct digit-wise base-10 representations?} 
Digit-wise representations may be more robust to noise in computations. 
If the number 120 is represented in value space, and has 1\% of relative noise introduced as a result of an operation, it may now be represented as 121 instead, leading to a mistake in the model's generation. Conversely, if 120 is represented in `digit space', an error of 1\% is not enough to change any of the digits independently. That is, the model can self-correct the number after the operation. 
Regarding the specific usage of base 10, 
one can presume it is because of the bias in the model's training data. That is, the model often has uses for the digits of a number, which biases the model toward learning to represent numbers in base 10, and as a result using that representation during operations.

\section*{Limitations}
Our experiments show that the digit-wise circular representations exist and can be extracted, and that they are more significant causally than previously described representations of magnitude and are sufficient for arithmetic. However, we do not show conclusively that the representation is the only representation of numeracy in the hidden representations of LLMs. That is, there may be a superposition of multiple redundant representations. Finally, our focus was exclusively on the natural numbers - which are only a subset of the numeric values that exist. Nevertheless, the natural numbers are the most prevalent and a natural starting point, and it could be expected that the digit-wise base 10 representation extends also to fractions, which we leave for future work to explore.

\section*{Acknowledgements}
We thank Amir Globerson and Daniela Gottesman for constructive feedback. This research was supported in part by Len Blavatnik and the Blavatnik Family foundation.

\bibliography{main}

\appendix

\section{Additional Results}

\subsection{Error patterns of GPT-4o}
\label{apx:gpt4o}
\begin{figure}[t]
\setlength\belowcaptionskip{-10pt}
\centering
   \includegraphics[scale=0.25]{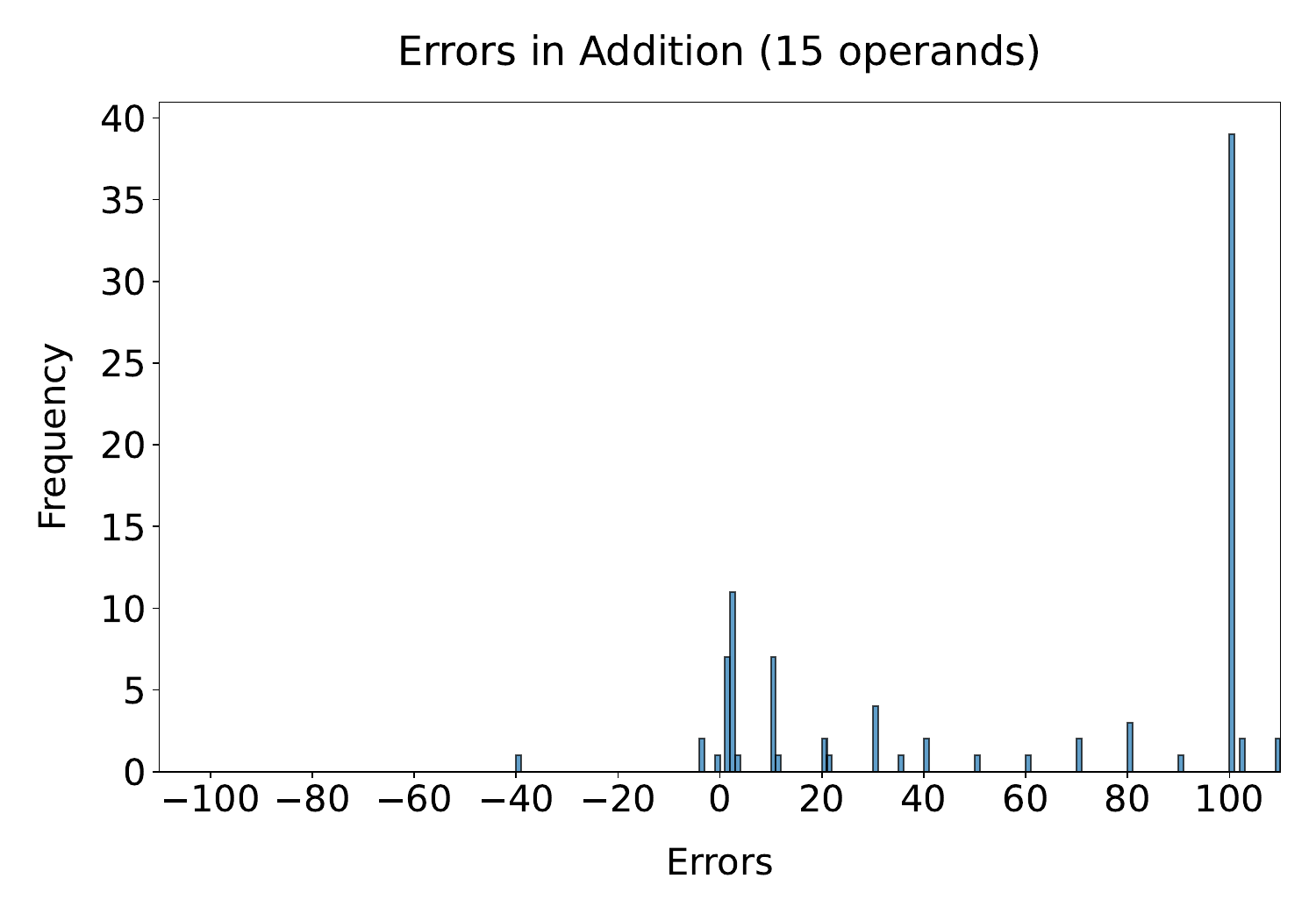}
  \caption{Error distribution in 15 operand addition for GPT-4o.}
    \label{fig:error_plotgpt4o}
\end{figure}

We conducted an additional error analysis using GPT-4o \citep{openai2024gpt4ocard} on 15-operand addition tasks. Increasing the number of operands was necessary due to the model's high accuracy on simpler addition problems. The results were consistent with the trends observed in Figure~\ref{fig:7operand}, showing the majority of errors are at multiples of 10, as seen in Figure~\ref{fig:error_plotgpt4o}. This indicates that the fragmented error distribution identified in smaller models persists in larger models.

Increasing the number of digits instead of the number of operands leads to errors in multiples of 100 and 1,000 as well, showing that the error distribution stays indicative of a fragmented representation also for other digits.

\subsection{PCA of hidden representations}
\label{apx:pca}

\begin{figure}[ht]
   \includegraphics[width=\columnwidth]{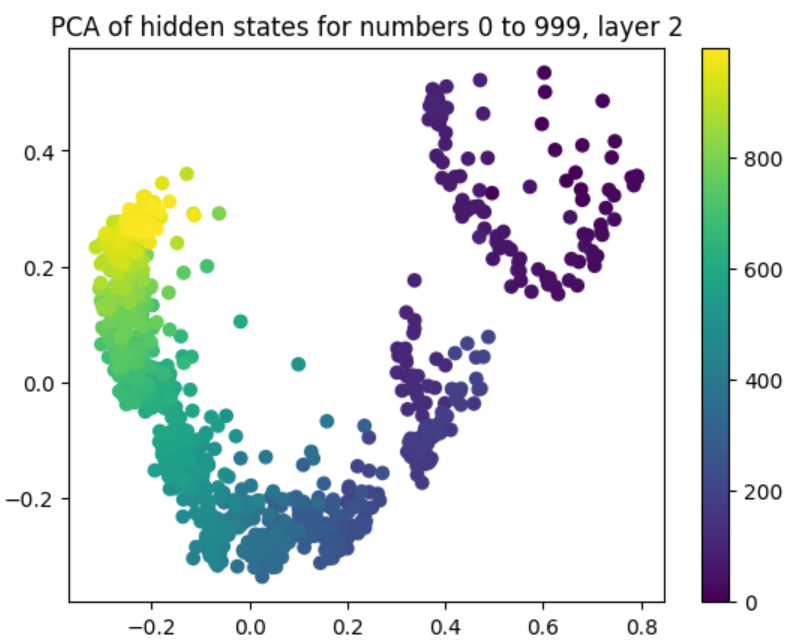}
  \caption{Visualization of the hidden states for natural number tokens (0 to 999) in layer 2 of Llama 3 8B, projected onto their top two principal components.}
    \label{fig:pca_all}
\end{figure}

We visualized the hidden states for natural number tokens 0 to 999 in layer 2 of Llama 3 8B, projected onto their top two principal components. In Figure~\ref{fig:pca_all} we can see that there are two half circles, one contained at the edge of the other. One is a half circle of all the numbers, and the next is of all numbers 0-99.

An interesting observation is that within each half-circle, the numbers increase in a clockwise direction, indicating that the model may represent digits circularly. In the circle for the numbers 0-99, the numbers increase clockwise, and again when you look at the half-circle that contains the rest of the numbers. This indicates that at least the hundreds digits and tens digits are represented circularly.

\begin{figure}[ht]

   \includegraphics[width=\columnwidth]{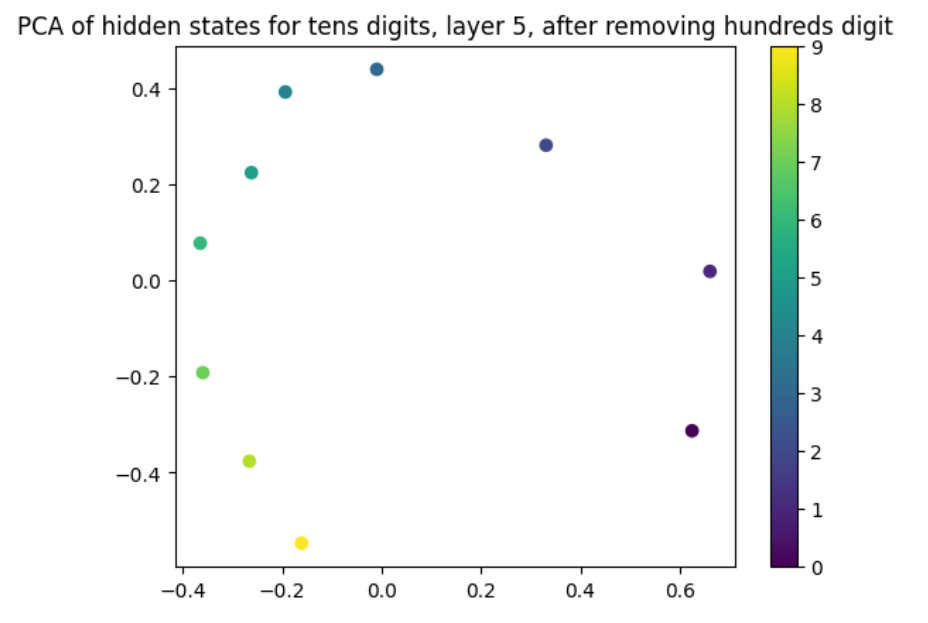}
  \caption{Visualization of the averaged hidden states for natural number tokens (0 to 999), grouped by their tens digit (0—9), in layer 5 of Llama 3 8B, projected onto their top two principal components. For example, numbers like 101 and 406, both having a tens digit of 0, are grouped together.}
      \label{fig:pca_tens_digit}
\end{figure}

In Figure~\ref{fig:pca_tens_digit} we can see that the circular pattern in the tens digit also extends to all numbers 0 to 999, when the dominance of the hundreds digit is removed through averaging out all numbers into 10 groups by their tens digit.

\subsection{Accuracy of circular probes}
\label{apx:probes}
\begin{figure}[ht]
\centering
    \includegraphics[scale=0.45]{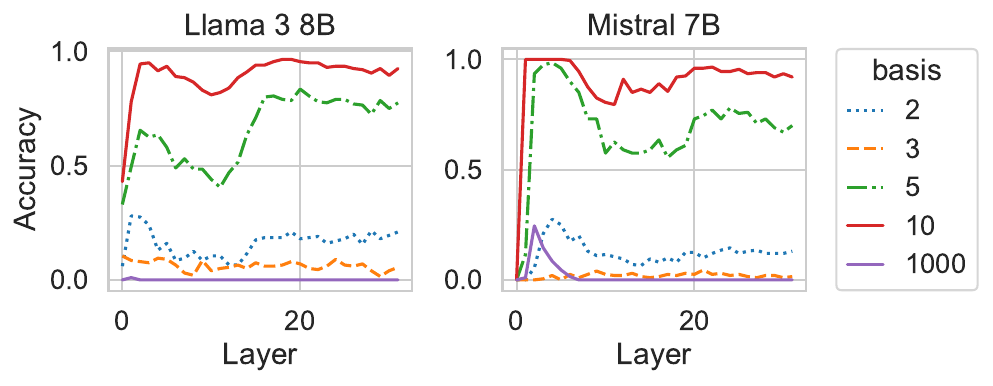}
    \caption{Accuracy of the circular probes in different bases across layers in Llama (left) and Mistral (right).}
    \label{fig:accuracy_per_layer}
\end{figure}

In the main results we showed the accuracy of the circular digit-wise probes, averaged over layers $\geq 3$. Here we will show this choice is justified as can be seen in Figure \ref{fig:accuracy_per_layer}. While there is significant variations between layers, the accuracy is especially low before the contextualization that happens in the first 3 layers.

Another interesting question is which layer's set of digit-wise circular-probes have the highest accuracy in predicting the number, and how accurate is it. The corresponding results can be seen in Table~\ref{table:max_accuracy}. It can be observed that in Mistral 7B, in the best layer, the circular probes achieve perfect accuracy on the validation set. That is, the number can always be recreated perfectly.

\begin{table*}[t]
\setlength\tabcolsep{4pt}
\footnotesize
\centering
\begin{tabular}{l*{15}{l}}
\toprule
Basis &  2    &  3    &  4    &  5    &  6    &  7    &  8    &  9    &  10   &  11   &  12   &  13   &  14   & 1000 & 2000 \\
\midrule
Llama 3 8B &  0.24 &  0.10 &  0.25 &  \underline{0.84} &  0.08 &  0.10 &  0.10 &  0.11 &  \textbf{0.96} &  0.12 &  0.09 &  0.10 &  0.11 & 0.00 & 0.02 \\
Mistral 7B &  0.28 &  0.04 &  0.22 &  \underline{0.98} &  0.04 & 0.08 &  0.12 &  0.23 &  \textbf{1.00} &  0.29 &  0.08 &  0.18 &  0.10 & 0.14 & 0.03 \\
\bottomrule
\end{tabular}
\caption{
Accuracy of the digit-wise circular probes for different bases in predicting all digits correctly, taking the layer with the highest accuracy.}
\label{table:max_accuracy}
\end{table*}

\subsection{Probing representations of numbers in word form}
\label{apx:wordform}
Since numbers can also be represented in word form, i.e. twenty-two for the number 22, we further tested if our digit-wise circular probes extend to these representations, without being explicitly trained on them. Concretely, we evaluated the probes' accuracy for Llama 3 8B on the numbers 'zero' through 'fifty' in word form.

We observe that the accuracy varies depending on the layer used, with a peak of 68.6 accuracy when using the representations at layer 14.
This is an encouraging sign that the circular probes generalize beyond the specific setting they were trained on, which further supports our causal results.

\section{Causal Intervention Details}
\label{apx:causaldetails}

We provide additional details on the interventions performed in \S\ref{sec:intervention}.
In practice, since the two directions of the circular probe are approximately orthogonal, we project the hidden representation onto each direction, subtract these components to remove the original digit representation, and then add the components back with their directions reversed to modify the digit. We also scaled the projection by a fixed constant ($a=19$), assuming that if the model has multiple representations for numbers, scaling the representation will make the model place more weight upon it. The exact constant was chosen through binary search, in order to select the largest scaling factor such that the model still predicts a number, as it was observed that with a very high scaling factor the model starts predicting non-numeric tokens.

\end{document}